\documentclass[letterpaper, 10 pt, conference]{ieeeconf}  

\IEEEoverridecommandlockouts                              

\usepackage{amsmath,amssymb,amsfonts}

\newtheorem{problem*}{Problem}

\usepackage[linesnumbered]{algorithm2e}
\usepackage{graphicx}
\usepackage{textcomp}
\usepackage{xcolor}
\usepackage{breqn}
\usepackage{tikz}
\usetikzlibrary{shapes,arrows}
\usepackage{bm,times}
\usetikzlibrary{arrows, decorations.markings}
\usepackage[noadjust]{cite}

\usepackage{etoolbox}
\makeatletter
\patchcmd{\@makecaption}
{\scshape}
{}
{}
{}
\makeatletter
\patchcmd{\@makecaption}
{\\}
{.\ }
{}
{}
\makeatother

\newcommand{\Tr}{\mathop{\bf Tr}}
\newcommand{\iou}{\mathop{\bf IoU}}
\newcommand{\distf}{\mathop{\bf d_{f}}}
\newcommand{\distn}{\mathop{\bf d_{v}}}
\newcommand{\diste}{\mathop{\bf d_{e}}}
\newcommand{\vecall}{\mathop{\bf vec}}

\tikzstyle{vecArrow} = [thick, decoration={markings,mark=at position
   1 with {\arrow[semithick]{open triangle 60}}},
   double distance=1.4pt, shorten >= 5.5pt,
   preaction = {decorate},
   postaction = {draw,line width=1.4pt, white,shorten >= 4.5pt}]
\tikzstyle{innerWhite} = [semithick, white,line width=1.4pt, shorten >= 4.5pt]

\usepackage{lastpage}
\usepackage{fancyhdr}

\usepackage{subfigure}

\title{\LARGE \bf SemanticLoop: loop closure with 3D semantic graph matching}

\author{Junfeng Yu, Shaojie Shen
\thanks{*The authors are with the Department of Electronic and Computer Engineering, Hong Kong University of Science and Technology, Hong Kong, China. (e-mail: jyubm@connect.ust.hk; eeshaojie@ust.hk).}
}

\begin{document}

\maketitle
\thispagestyle{empty}
\pagestyle{empty}


\begin{abstract}
    Loop closure can effectively correct the accumulated error in robot localization, which plays a critical role in the long-term navigation of the robot. Traditional appearance-based methods rely on local features and are prone to failure in ambiguous environments. On the other hand, object recognition can infer objects' category, pose, and extent. These objects can serve as stable semantic landmarks for viewpoint-independent and non-ambiguous loop closure. However, there is a critical object-level data association problem due to the lack of efficient and robust algorithms.

    We introduce a novel object-level data association algorithm, which incorporates IoU, instance-level embedding, and detection uncertainty, formulated as a linear assignment problem. Then, we model the objects as TSDF volumes and represent the environment as a 3D graph with semantics and topology.
    Next, we propose a graph matching-based loop detection based on the reconstructed 3D semantic graphs and correct the accumulated error by aligning the matched objects. Finally, we refine the object poses and camera trajectory in an object-level pose graph optimization.

    Experimental results show that the proposed object-level data association method significantly outperforms the commonly used nearest neighbor method in accuracy. Our graph matching-based loop closure is more robust to environmental appearance changes than existing appearance-based methods.
\end{abstract}


\section{Introduction}\label{sec:Introduction}

The long-term autonomous navigation of mobile robots is critical for many applications (e.g., self-driving cars and service robots). However, accumulated errors will inevitably occur in robot localization due to sensor noise. In order to correct the accumulated drift, robots need to perceive the environment in real-time and recognize previously visited places (i.e., loop closure). Although loop closure has been studied extensively in Simultaneous Localization and Mapping (SLAM), it is still considered a well-defined but highly challenging problem to solve in the general sense.

Classical appearance-based methods typically reformulate loop closure as an image retrieval problem. They represent the environment as a database of images. Then the current image is matched with the ones in the database to retrieve the most similar candidate(s) in appearance. These methods generally use visual descriptors to represent images for more efficient retrieval. The Bag-of-Words (BoW~\cite{sivic2003video}) extracted from local features (e.g., ORB~\cite{rublee2011orb}) is one of the most effective models. Many existing SLAM systems (e.g., ORB-SLAM2~\cite{mur2017orb}, VINS-Mono~\cite{qin2018vins}) used BoW and demonstrated impressive performance. These approaches are flexible and general. However, they still face many challenges. For example, when the appearance changes due to lighting or viewpoint differences, the local features may change dramatically, and the classical methods fail. Moreover, appearance-based methods tend to ignore the geometric structure of the environment, which may lead to false positives in repetitive environments.

On the other hand, semantics and geometric structures are usually invariant to appearance changes. For example, a chair remains a chair, whether observed during the day or night or from different viewpoints. Recently, deep learning has made significant progress on perceptual tasks such as object detection and instance segmentation (e.g., Mask R-CNN~\cite{he2017mask}), motivating the incorporation of semantics into SLAM systems to improve the localization accuracy (e.g., SLAM\texttt{++}~\cite{salas2013slam++}, Fusion\texttt{++}~\cite{mccormac2018fusion++}). However, due to the generalization problem, deep learning models often suffer from noise (e.g., false detections, misclassifications) in the working environment. This perceptual noise can easily lead to incorrect object-level data associations, which introduces erroneous semantics. Although this inaccurate semantic information can seriously affect the accuracy and robustness of localization, existing works tend to ignore this critical problem.

\paragraph*{Contributions} To address the above challenges, we propose an RGBD-based semantic mapping system with loop closure. Specifically, our main contributions are as follows:
\begin{itemize}
    \item We introduce a novel object-level data association method that combines IoU, instance-level embedding, and detection uncertainty into a linear assignment formulation, constructing an accurate 3D semantic map insensitive to the noises from deep learning models and odometry drift.
    \item We propose a 3D semantic graph matching-based loop closure approach that couples semantics and topology of the instances in a quadratic assignment formulation, making the loop closure more robust to appearance changes in the scene.
    \item To maintain a globally consistent map, we introduce an object-level pose graph optimization that includes the odometry and loop closure constraints to optimize the camera trajectories and object poses jointly.
\end{itemize}
Moreover, we evaluate the proposed methods on the public TUM RGBD benchmark~\cite{sturm12iros} and SceneNN dataset~\cite{hua2016scenenn} to verify their effectiveness.


\section{Related Work}\label{sec:Related_Work}
\subsection{Data Association in Semantic SLAM}\label{subsec:Data_Association_in_Semantic_SLAM}

SLAM\texttt{++}~\cite{salas2013slam++} is a pioneering work in the direction of semantic SLAM, which first used real-world objects (e.g., tables, chairs) as landmarks. The data association relies on the Point-Pair feature (PPF)-based 3D object recognition. However, it required a pre-built database of CAD models, making the system less universal. In their later work, Fusion\texttt{++}~\cite{mccormac2018fusion++} utilized a reconstruction-by-segmentation strategy to build a TSDF volume for each object, which solved the problem of relying on an offline CAD database. The data association depended on the IoU between the mask from instance segmentation and the mask projected from the TSDF volume. However, the data associations can become ambiguous when odometry drifts or objects are occluded.

Another line of study is the probabilistic data association. The core idea is to use 'soft' instead of 'hard' data association to put the data association uncertainty into the SLAM backend. The most representative work is Gaussian PDA~\cite{bowman2017probabilistic}, which proposed using the Expectation-Maximization (EM) algorithm to solve this discrete-continuous optimization problem. The EM algorithm needs to be solved iteratively. However, recalculating the combinatorial number of historical data associations is infeasible for computational reasons. Semantic MM~\cite{doherty2020probabilistic} advanced this stream by approximating the data associations with a Max-Marginalization (MM) technique, which solved the computational complexity problem by assuming that future observations will not affect past data associations. However, not optimizing past data association weights may result in a low probability of getting the correct data association, especially when there are many ambiguities due to odometry drift.

Recently, QuadricSLAM~\cite{nicholson2018quadricslam} and CubeSLAM~\cite{yang2019cubeslam} explored the use of ellipsoids and cuboids as object representations, which were extracted from multi-view and single-view 2D bounding boxes, respectively. In QuadricSLAM, they overlooked the data association problem and focused on ellipsoid initialization. In the following work,~\cite{qian2021semantic} used the BoW model as object representation and formulated the data association as a linear assignment problem. In CubeSLAM, the data association relied on feature point matching, and the bounding box that shared the most feature points was selected. However, data association in these works depended on traditional feature points or descriptors without semantics, which may be subject to failure on textureless objects (e.g., TV).

Similar to Fusion\texttt{++}~\cite{mccormac2018fusion++}, we use a reconstruction-by-segmentation strategy. The dense object model can represent objects' pose and shape while providing sufficient semantics. In contrast to the commonly used nearest neighbor method, our data association method combines IoU, instance-level embedding, and detection uncertainty into a linear assignment formulation. Therefore,  it is more robust to textureless objects, deep learning model noises, and odometry drift.


\subsection{Loop Closure in Semantic SLAM}\label{subsec:Loop_Closure_in_Semantic_SLAM}

SLAM\texttt{++}~\cite{salas2013slam++} performed loop closure by matching the local object graph with the long-term object graph. They treated objects as vertices and their x-axes as normal directions to extract the PPFs and then reused the same 3D object recognition algorithm as in data association. In Fusion\texttt{++}~\cite{mccormac2018fusion++}, they extracted 3D BRISK for object models and applied the 3D-3D RANSAC algorithm between them to perform loop detection, which was extremely slow (more than 780ms) even on modern GPU platforms.

Recent approaches attempted to incorporate more semantics to address the loop closure problem in cases with extreme appearance changes. X-View~\cite{gawel2018x} proposed a novel loop detection idea based on semantic graphs. The system constructed 2D semantic graphs using image sequences with instance segmentation, in which vertices were semantic blobs and edges represented proximity relations. Loop closure depended on matching the random walk descriptors between vertices. The random walk descriptor contained topological information of the semantic graph, making it highly robust to seasonal and significant viewpoint changes.
A series of follow-up works had extended the idea in X-View to 3D~\cite{liu2019global}, to edit distance minimization-based matching algorithm~\cite{lin2021topology}, and to semantic histogram-based descriptor~\cite{guo2021semantic} (to be faster).
However, when there are many objects in the graph, random walks tend to lose information, and the performance may degrade severely.
Graph matching in~\cite{lin2021topology} often suffers falses alarms when duplicated objects with similar topologies are in the graph.
Moreover, matching descriptors between the query and target graphs is inefficient when the number of random walks is large.

In our system, we perform loop closure by constructing 3D semantic maps online. To improve the efficiency and robustness of loop closure detection, we perform a geometric graph matching between semantic graphs rather than descriptors matching between vertices. In addition, our system is a complete pipeline, including a pose graph optimization to maintain the camera and object poses.


\section{Method}\label{sec:method}
\subsection{Overview}\label{subsec:Overview}

\begin{figure*}[htb]
    \centering
    \begin{minipage}[b]{0.85\linewidth}
        \includegraphics[width=0.85\linewidth]{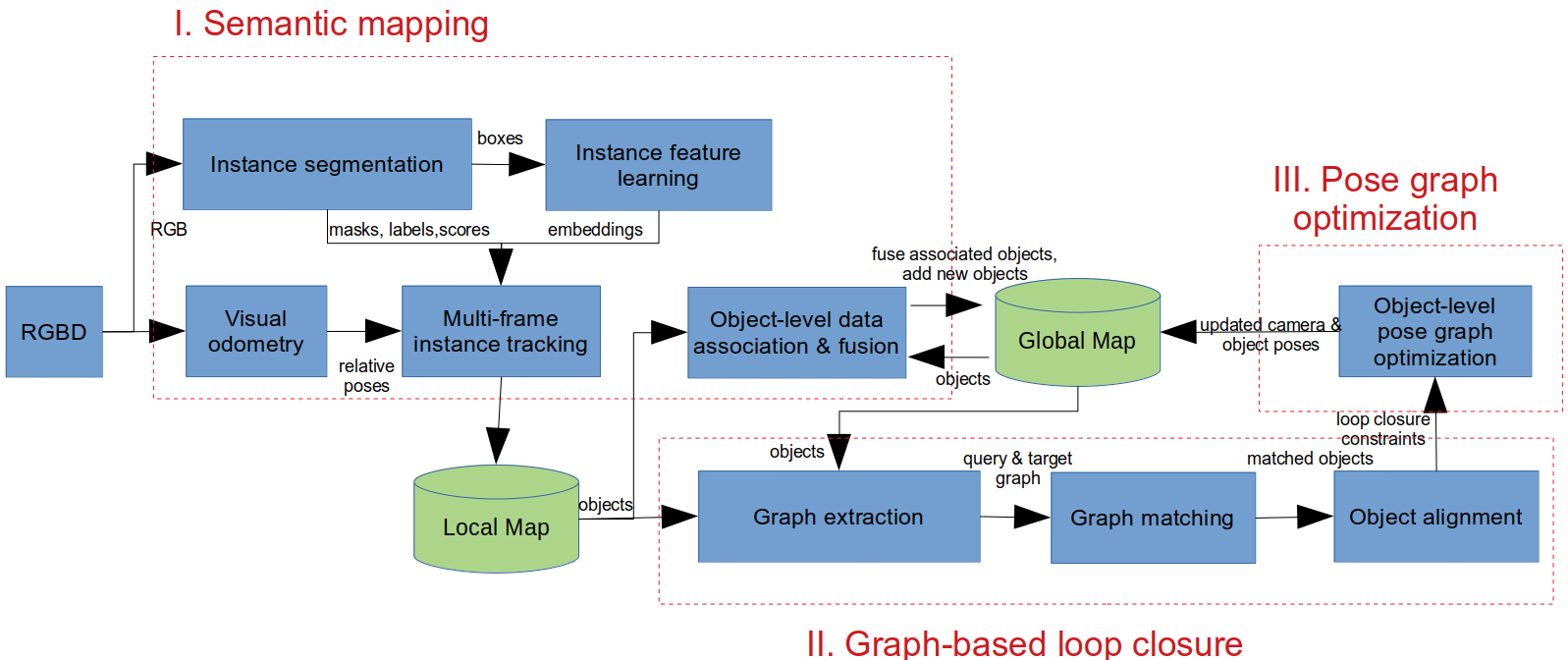}
    \end{minipage}
    \caption{Overview of the proposed semantic mapping system with loop closure.}
    \label{fig:pipeline}
\end{figure*}

Figure~\ref{fig:pipeline} visualizes the pipeline proposed in our work. There are mainly three modules: semantic mapping, graph-based loop closure, and pose graph optimization.

\subsection{Semantic Mapping}\label{subsec:Semantic_Mapping}
From RGBD input, we utilize off-the-shelf RGBD odometry to obtain relative poses. The underlying assumption is that we can build a 3D object map based on the local consistency of the odometry and our data association algorithm.
An instance segmentation network processes the RGB frame in a separate thread to detect bounding boxes, masks, and semantic labels.
Then the bounding boxes are fed into an instance-feature learning network to extract the instance-level embeddings.
Based on the outputs of the neural network thread, the multi-frame instance tracking algorithm filter out perceptual noises and integrate a local map at the current position.
Then, the object-level data association algorithm matches detections in the local map with the objects in the global map based on semantics and camera poses.
When no match occurs, we create a new TSDF volume and add it to the global map.
When an object is associated, we utilize an approach similar to Fusion\texttt{++}~\cite{mccormac2018fusion++} to fuse the new measurements into the TSDF volume and use an averaging scheme to update a probability distribution over the semantic label.

\paragraph*{Object-Level Data Association}
The core of semantic mapping is a critical object-level data association problem.
Suppose at frame $i$, there are $M$ detections from the instance segmentation network, denoted as $\mathcal{S} \triangleq \left \{ s_{k} \right \}_{k=1}^{M}$. Each detection is represented as $s_{k} = \left ( m_{k}, l_{k}, c_{k}, e_{k} \right )$, where $m_{k}$ is the binary mask, $l_{k}$ is the semantic label, $c_{k}$ is the confidence score, and $e_{k}$ is the embedding from the instance feature learning network.
Meanwhile, we have $N$ object landmarks in the object map, denoted as $\mathcal{O} \triangleq \left \{ o_{j} \right \}_{j=1}^{N} $. Each object is represented as $o_{j} = \left ( V_{j}, T_{wo_{j}}, l_{j}, m_{j}, E_{j} \right )$, where $V_{j} $ is the TSDF volume, $T_{wo_{j}}$ is the pose, $l_{j}$ is the semantic label, $m_{j}$ is the predicted binary mask, and $E_{j}$ is a set storing all the matched embeddings from past matches.

The object-level data association needs to find as many matches as possible by assigning at most one object to each detection and at most one detection to each object, such that the total matching cost is minimized. Since we usually have more landmarks than detections, i.e., N $\ge$ M, we can reformulate this problem as a 2D rectangular assignment problem as follows:


\begin{align}
    \begin{split}
        \min_{\mathbf{A}} & \sum_{j=1}^{N} \sum_{k=1}^{M} \Tr(\mathbf{A}^{\top} \mathbf{L}) \\
        \text{subject to } & \mathbf{A}\left ( j,k \right ) \in \left \{ 0,1 \right \} ,\forall j, k\\
        &\sum_{j=1}^{N} \mathbf{A}\left ( j,k \right ) = 1 ,\forall k \\
        & \sum_{k=1}^{M} \mathbf{A}\left ( j,k \right ) \le 1 ,\forall j \\
    \end{split}
    \label{eqn:linear_assignment_formulation}
\end{align}
where $\mathbf{A}$ is a $N \times M$ assignment matrix, and $\mathbf{L}$ is a $N \times M$ cost matrix. The equality constraint means that every column (detection) is assigned to a row (landmark). The inequality constraint means that not every row (landmark) is assigned to a column (detection).
Note that due to unobserved new objects or false detections, the matching cost between a detection and a landmark may surpass a certain threshold, then this association should be discarded.

The assignment matrix $\mathbf{A}$ can be defined as follows:
\begin{equation}
    \mathbf{A}\left ( j,k \right )
    =
    \begin{cases}
         & 1\text{, if }o_{j} \text{ is matched with } s_{k} \\
         & 0\text{, otherwise }
    \end{cases}
    \label{eqn:data_association_assignment_matrix}
\end{equation}

A matching cost between object $o_{j}$ and detection $s_{k}$ can be calculated based on the binary masks, embeddings, and semantic labels as follows:
\begin{align}
    \begin{split}
        &\mathbf{L}\left ( j,k \right )
        =
        1.0 - \mathbf{W}\left ( j,k \right ) p \left ( l_{k} | l_{j}\right ) \\
        &\mathbf{W}\left ( j,k \right )
        =
        \lambda \iou \left ( j, k \right ) + \left ( 1.0 - \lambda \right ) \distf \left ( j, k \right )\\
        &\iou \left ( m_{j}, m_{k} \right )
        =
        \frac{\sum m_{k} \cap m_{j}}{\sum m_{k} + \sum m_{k} - \sum m_{k} \cap m_{j}} \\
        &\distf \left ( e_{j}, e_{k} \right )
        =
        e_{j} e_{k}
    \end{split}
    \label{eqn:data_association_cost_matrix}
\end{align}
where $\iou$ is calculated between the mask $m_{k}$ and the predicted mask $m_{j}$.
The metric distance ($\distf$) of embeddings is computed between the instance embedding $e_{k}$ and every embedding $e_{j} \in E_{j}$.
We use the cosine distance and choose the maximum among all $\distf \left ( e_{j}, e_{k} \right )$.
A hyperparameter $\lambda$ is used to balance the $\iou$ and $\distf$ terms.
The probability distribution $p \left ( l_{k} | l_{j}\right ) $ corresponds to the confusion matrix of the instance segmentation network and is learned offline.
The problem defined in equation $\left (\ref{eqn:linear_assignment_formulation} \right )$ can be solved using the shortest augmenting path algorithm, described in~\cite{crouse2016implementing}.

\paragraph*{Multi-Frame Instance Tracking}
Since noises of the deep learning model are likely to lead to erroneous data associations, we perform a multi-frame instance tracking to filter out the perceptual noises. We reuse the same formulation as in object-level data association, and the main difference is to use the mask in the previous frame instead of the predicted mask. We only keep the instances that are tracked over a certain number of times.

\subsection{Graph-based Loop Closure}\label{subsec:Graph-based_Loop_Closure}
\begin{figure}[htb]
    \begin{center}
        \includegraphics[width=0.9\columnwidth]{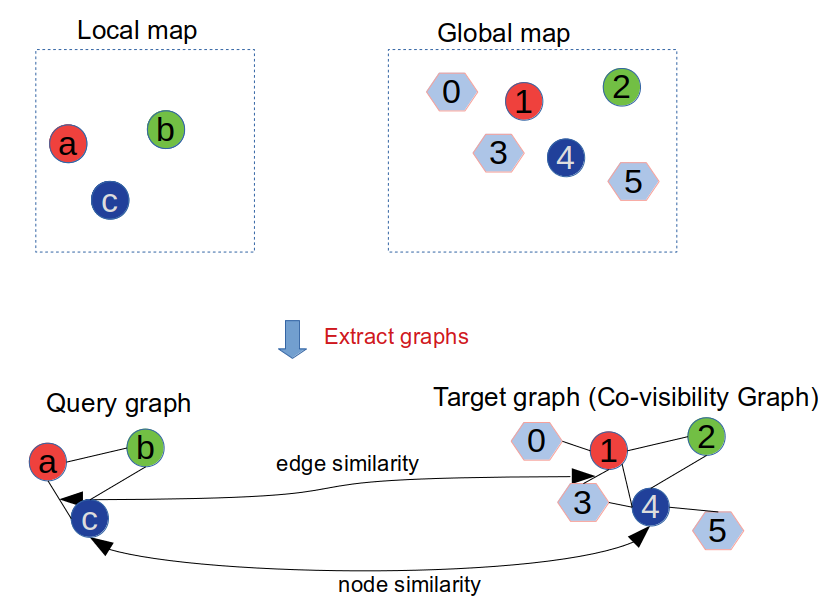}
    \end{center}
    \caption{An example shows the graph extraction and graph matching steps.
        The object in the map forms a graph, where the vertex is composed of the object's center and semantics. The edge represents the metric distance and co-visibility relationship between the objects. Graph matching aims to find the correspondences between the query and target graphs by considering both edge and vertex similarity.}
    \label{fig:graph_extraction_example}
\end{figure}

Directly matching between vertices through random walk descriptions is inefficient and does not fully exploit the topology in the graph.
First, we extract the query and target graphs from the local and global maps.
Then, we perform a geometric graph matching to find the correspondences between the query and target graphs by considering both edge and vertex similarity.
Next, we estimate the drift errors by aligning the matched objects.
See an example in Figure~\ref{fig:graph_extraction_example}.

\paragraph*{Graph Extraction}
All objects in the map form a graph $G=\left ( V, E \right ) $, where vertices $v_{j} \in V$ contains the center, semantic label $l_{j}$ and embeddings of object $o_{j}$. Each edge $e_{j1,j2} \in E$ is given by the Euclidean distance between the centers of object $o_{j1}$ and object $o_{j2}$. In order to maintain the topology between the objects, we adopt a co-visibility strategy. That is, we add an edge between the nodes only when the corresponding objects are observed in the same local map. In this way, we extract the query $G_{q}$ and target $G_{t}$ graphs from the local and global maps.

\paragraph*{Graph Matching}
With a query graph $G_{q}=\left ( V_{q}, E_{q} \right )$ of size $M$ and a target graph $G_{t}=\left ( V_{t}, E_{t} \right )$ of size $N$, the graph matching step aims to find a correspondence between graphs, which fits both vertex's attributes (e.g., semantic label, embeddings) and graph topology (e.g., Euclidean distance between the co-visible objects). The problem can be reformulated as a quadratic assignment problem as follows:
\begin{align}
    \begin{split}
        \max_{\mathbf{A}} &\sum_{e_{j1,j2} \in E_{q}} \sum_{ e_{k1,k2} \in E_{t} }
        (
        \mathbf{A}\left ( j_{1},k_{1} \right )
        \mathbf{A}\left ( j_{2},k_{2} \right ) \\
        &\mathbf{L}\left ( j_{1}, j_{2}, k_{1}, k_{2} \right ) )\\
        =\max_{\mathbf{A}}
        & \vecall \left ( \mathbf{A} \right )^\top \mathbf{S} \vecall \left ( \mathbf{A} \right )\\
        \text{subject to } &\mathbf{A}\left ( j,k \right ) \in \left \{ 0,1 \right \} ,\forall j, k\\
        & \sum_{j=1}^{N} \mathbf{A}\left ( j,k \right ) \le 1 ,\forall k \\
        & \sum_{k=1}^{M} \mathbf{A}\left ( j,k \right ) \le 1 ,\forall j \\
    \end{split}
    \label{eqn:quadratic_assignment_formulation}
\end{align}
where $\mathbf{A}$ is a $N \times M$ assignment matrix,
and $\mathbf{L}$ is a $N \times M \times N \times M$ reward tensor,
The $\vecall$ operator vectorizes a matrix into a column vector.
The reward matrix $\mathbf{S}$ is a square matrix of size $N M$, which is constructed by unfolding the reward tensor $L$.
The diagonal elements of the reward matrix are the matching reward for the nodes, and the off-diagonal elements are the matching reward for the edges.

The assignment matrix $\mathbf{A}$ can be defined as follows:
\begin{equation}
    \mathbf{A}\left ( j,k \right )
    =
    \begin{cases}
         & 1\text{, if }v_{j} \in V_{t} \text{ is matched with } v_{k} \in V_{q} \\
         & 0\text{, otherwise }
    \end{cases}
    \label{eqn:graph_matching_assignment_matrix}
\end{equation}

A matching reward can be defined as follows:
\begin{align}
    \begin{split}
        &\mathbf{L}\left ( j_{1}, j_{2}, k_{1}, k_{2} \right )
        =
        \distn \left ( j_{1}, k_{1} \right )
        \distn \left ( j_{2}, k_{2} \right )
        \diste \left ( e_{j1,j2}, e_{k1,k2} \right )\\
        &
        \distn \left ( j, k \right )
        =
        \begin{cases}
             & 1 \text{ , if } l_{j} = l_{k} \\
             & 0 \text{ , otherwise }
        \end{cases} \\
        &\diste \left ( e_{j1,j2}, e_{k1,k2} \right )
        =
        \exp \left ( -\mu \left \| e_{j1,j2} - e_{k1,k2} \right \|_2  \right )
    \end{split}
    \label{eqn:graph_matching_reward_matrix}
\end{align}
where functions $\distn$ and $\diste$ calculate the similarity of vertices and edges respectively, and $\mu$ is a hyperparameter.

The problem defined in equation $\left ( \ref{eqn:quadratic_assignment_formulation} \right ) $ is NP-hard. However, we can solve it approximately using the spectral methods described in \cite{leordeanu2005spectral}.
We first relax the integral constraints on $A$, such that the elements of $A$ can take real values between [0, 1].
Since only the relative values between the elements of $A$ matter, we can fix the norm of $\vecall(A)$ to 1.
According to the Raleigh's ratio theorem, the $\vecall(A^*)$ that maximizes $\vecall(A)^T S \vecall(A)$ is the principal eigenvector of $S$.
Moreover, a key constraint of $S$ is that it is element-wise non-negative.
Therefore, by the Perron-Frobenius theorem, the elements of $\vecall(A^*)$ are non-negative, i.e., between [0, 1].
Next, in order to obtain an assignment matrix from $A^*$, i.e., a matrix with elements in $\left \{ 0,1 \right \}$ and proper row/column sums, \cite{leordeanu2005spectral} proposed to use a greedy algorithm to discretize the $\vecall(A^*)$.
Our main difference from \cite{leordeanu2005spectral} is that we reuse the linear assignment formulation in equation $\left (\ref{eqn:linear_assignment_formulation} \right )$ with a cost matrix $A^*$ to obtain an assignment matrix.

\paragraph*{Object Alignment}
We estimate the relative transformations by registering the point clouds extracted from the TSDFs of the matched objects. In order to get an accurate result for this wide baseline alignment, we first use the FPFH~\cite{rusu2009fast}-based 3D-3D RANSAC to perform an initial coarse alignment, then use ICP for refinement.

\subsection{Pose Graph Optimization}\label{subsec:Pose_graph_Optimization}
\begin{figure}[htb]
    \begin{center}
        \includegraphics[width=0.9\columnwidth]{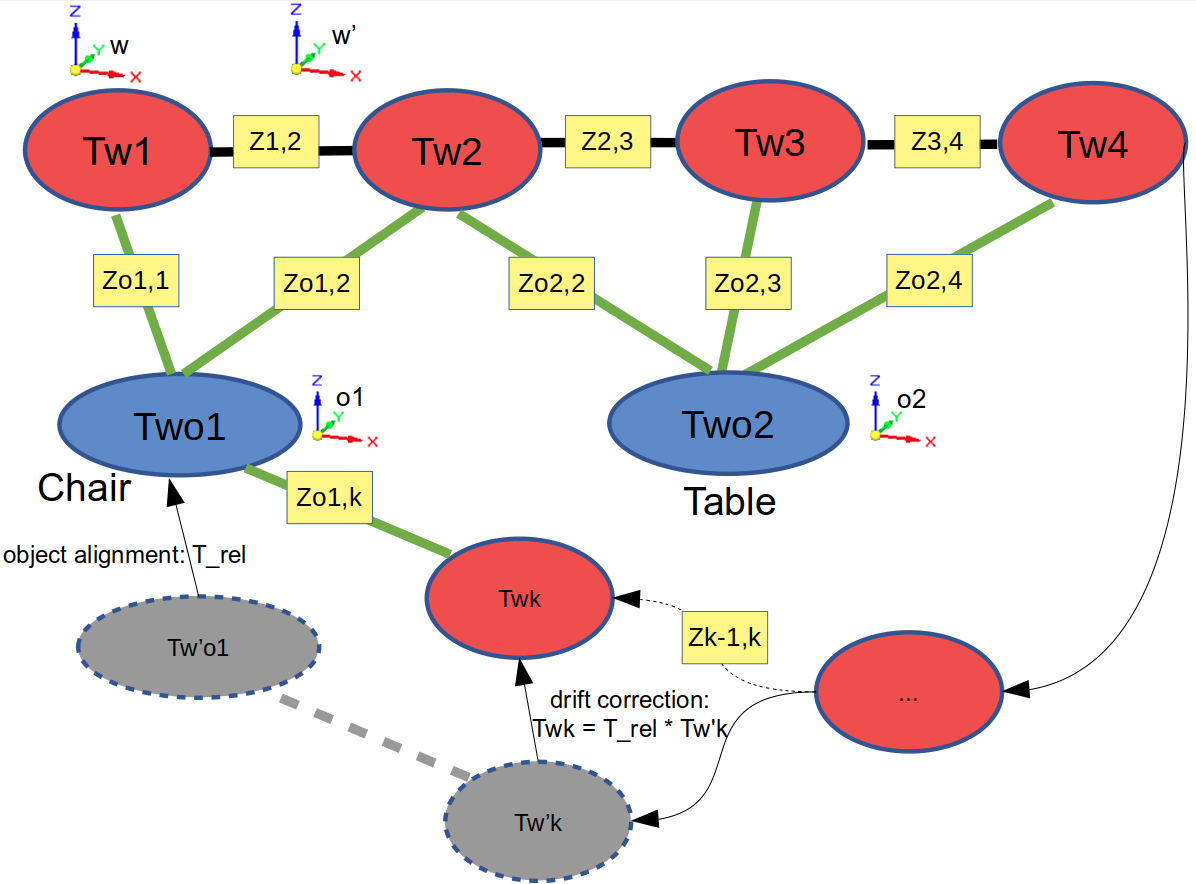}
    \end{center}
    \caption{An example pose graph shows six camera pose nodes $T_{wi}$ (red circles) and two object pose nodes $T_{wo_{j}}$ (blue circles).The object-camera constraints $Z_{o_{j}, i}$ (green edges) represent the pose of frame $i$ in the coordinate system of the object $o_{j}$. The camera-camera constraints $Z_{i,i+1}$ (black edges) denote the odometry measurements between frame $i$ and $i+1$. When a loop is detected, the drift error $T_{\_rel}$ is obtained by aligning the object in the local map (node $T_{w'o_{1}}$) with the object in the global map (node $T_{wo_{1}}$). Then, the camera pose $T_{wk}$  and object-camera constraint $Z_{o_{1},k}$ are corrected using the estimated drift error.}
    \label{fig:pose_graph_example}
\end{figure}

The pose graph contains both object and camera nodes. Each node contains an $\mathbf{SE(3)}$ transformation. For frame $i$ with instance segmentation, we create a new camera node $T_{wi}$. We fix the first camera node as the world coordinate system. When a new object $o_{j}$ is added, we create a corresponding object node $T_{wo_{j}}$. The object's coordinate system is attached to the object's center, and the coordinate axes are aligned with the world coordinate axes. Each $\mathbf{SE(3)}$ measurement is a relative transformation constraint between the corresponding nodes. The measurement $Z_{i,i+1}$ between camera nodes represents the relative pose estimate from frame $i$ to $i+1$. The measurement $Z_{o_{j}, i}$ between the object and camera nodes denotes the pose of frame $i$ expressed in the object $o_{j}$'s coordinate system.
See an example in Figure~\ref{fig:pose_graph_example}.

\paragraph*{Object-Level Pose Graph Optimization}
After getting the drift errors through object alignment, we can calculate the corrected camera pose and add the new object-camera constraints to the pose graph. Then we can further refine the entire camera trajectory $ \mathcal{X} = \left \{ T_{w,i} \right \}_{i=1}^{T} $ and object poses $ \mathcal{O} = \left \{ T_{w,o_{j}} \right \}_{j=1}^{M} $ in a object-level  pose graph optimization. We minimize the error terms for all measurement constraints as follows:
\begin{align}
    \begin{split}
        &\mathcal{X}, \mathcal{O}
        =
        \mathop{\arg\min}\limits_{\mathcal{X}, \mathcal{O}}
        \sum_{Z_{i,i+1}}{
        \left \|
        e_{cc} \left ( T_{w,i}, T_{w,i+1} \right )
        \right \|_{\Sigma_{t,t+1}}} \\
        &+
        \sum_{Z_{o_{j},i}}{
        \left \|
        e_{oc} \left ( T_{w,i}, T_{w,o_{j}} \right )
        \right \|_{\Sigma_{o_{j},i}}} \\
        &e_{cc} \left ( T_{w,i}, T_{w,i+1} \right )
        =
        \log_{}{(
            Z_{i,i+1}^{-1}}
        T_{w,i}^{-1}
        T_{w,i+1}) \\
        &e_{oc} \left ( T_{w,i}, T_{w,o_{j}} \right )
        =
        \log_{}{(
            Z_{o_{j},i}^{-1}}
        T_{w,o_{j}}^{-1}
        T_{w,i})
    \end{split}
    \label{eqn:pose_graph_optimization}
\end{align}
where $e_{cc}$ and $e_{oc}$ are the meansurement error terms for the camera-camera $Z_{i,i+1}$ and object-camera measurement constraint $Z_{o_{j},i}$ respectively.
$
    \left \|
    e
    \right \|_{\Sigma} = e^{\top} \Sigma^{-1} e
$
is the Mahalanobis distance and $\log$ is the logarithmic map of $\mathbf{SE}\left ( 3 \right ) $. We solve this nonlinear least squares problem using the Levenberg-Marquart algorithm in the Ceres solver. After optimization, we update the object and camera poses before initializing new objects.

\section{Experiments}\label{sec:Experiments}
We use the open-source Mask R-CNN implementation of Matterport as the instance segmentation network, and the weights are pre-trained on the Microsoft COCO dataset~\cite{lin2014microsoft}. We use~\cite{Wojke2018deep} as the instance-level feature learning network, and the weights are fine-tuned on the SceneNet RGBD dataset~\cite{mccormac2017scenenet}.

We conduct experiments on the public TUM RGBD and SceneNN dataset to evaluate the performance of the proposed methods. The TUM RGBD dataset~\cite{sturm12iros} consists of real-time RGB, depth images, and ground-truth trajectories. In addition to RGB, depth, and camera pose ground-truth, the SceneNN dataset~\cite{hua2016scenenn} provides instance segmentation ground-truth.

\subsection{Data Association Performance}\label{subsec:Data_Association_Performance}
\paragraph*{Metric}
The idea is to find the correspondences between ground-truth object IDs and object IDs in the data association algorithm. Similar to~\cite{qian2021semantic}, we reformulate this problem as a linear assignment problem. The first partite set $S$ consists of the object IDs in the data association algorithm. The second partite set $F$ consists of the ground-truth object IDs. The reward function $w\left ( j, k \right ) $ on edge $e\left ( j, k \right ) \in F \times S$ can be defined as the number of identical bounding boxes shared by the object ID $j \in F $ and object ID $k \in S$. By solving this problem, the sum of the reward on all matched edges is the number of correct associations. We take the ratio of the number of correct associations to the number of all ground-truth bounding boxes as the accuracy of the data association algorithm.

\begin{figure*}[htb]
    \centering
    \subfigure[Translational noise]{
        \begin{minipage}[b]{0.90\linewidth}
            \includegraphics[width=0.3\linewidth]{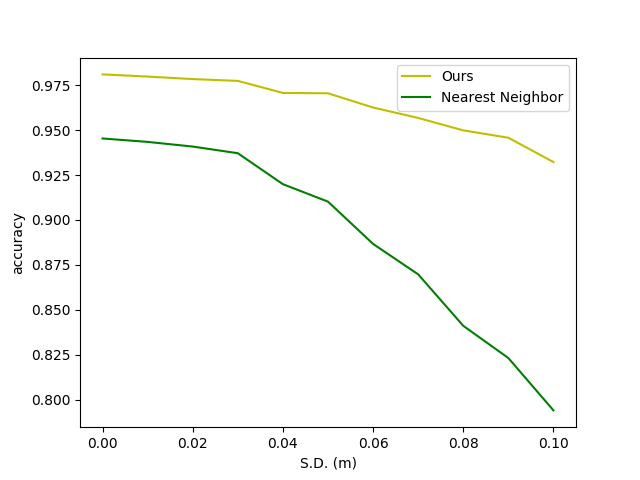}
            \includegraphics[width=0.3\linewidth]{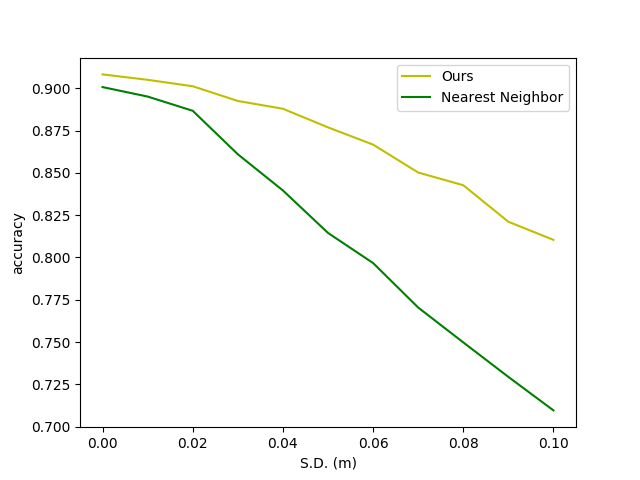}
            \includegraphics[width=0.3\linewidth]{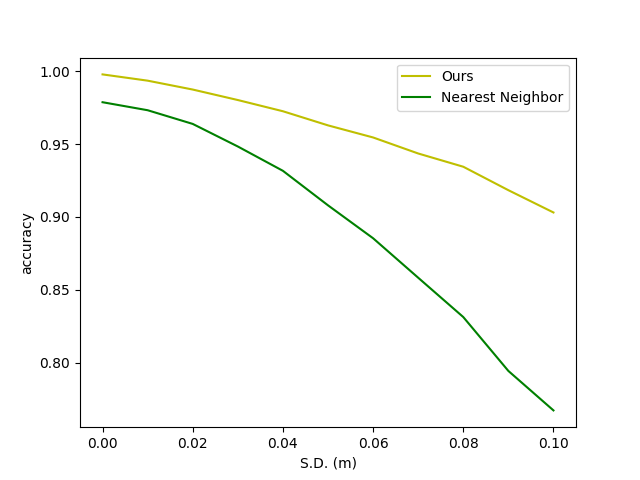}
        \end{minipage}}
    \subfigure[Orientational noise]{
        \begin{minipage}[b]{0.90\linewidth}
            \includegraphics[width=0.3\linewidth]{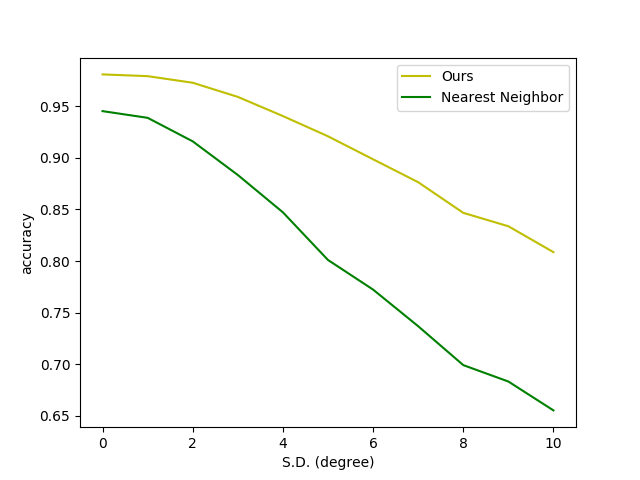}
            \includegraphics[width=0.3\linewidth]{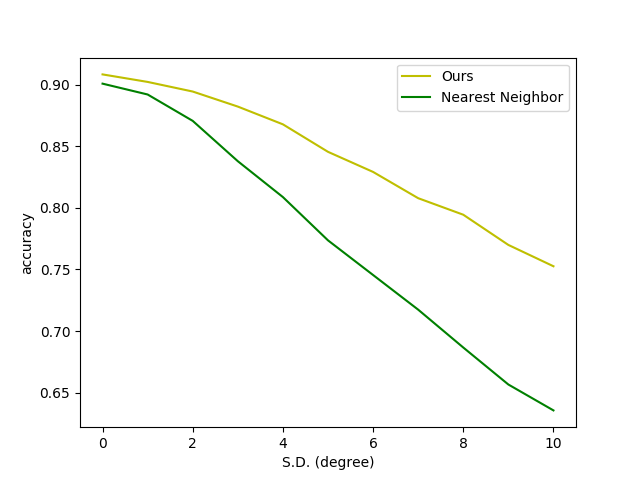}
            \includegraphics[width=0.3\linewidth]{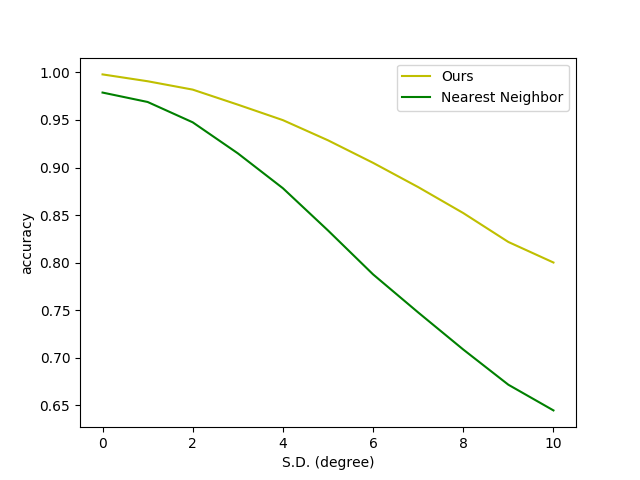}
        \end{minipage}}
    \caption{Results of object-level data association accuracy according to the different noise levels on the public SceneNN dataset. The three columns are 021, 025, and 231 sequences. (a) accuracy at different translational noise levels. (b) accuracy at different orientational noise levels.}
    \label{fig:data_association_results}
\end{figure*}

We compare the proposed method with the commonly used nearest neighbor method in semantic SLAM systems.
In our experiments, we add random noises to the ground-truth camera poses and observe how the accuracy changes with different noise levels. We conduct experiments on SceneNN 021, 025, and 231 sequences, as shown in Figure~\ref{fig:data_association_results}.
Due to the inaccuracy of the camera poses, it is easy to cause the prediction to deviate from the measurement, resulting in ambiguities in the data association. Results show that the proposed method is more robust to localization noise than the baseline method due to incorporating instance-level embeddings and performing global minimum cost matching.

\subsection{Loop Detection Performance}\label{subsec:Loop_Detection_Performance}
We label two frames as a loop closure if they observe more than two objects in common. To prevent adjacent frames from being labeled as loop closures, the difference in frame indices between them needs to be greater than 500. If 50\% of the edges in the local graph can be matched, then we consider the loop detection successful.
We compare the proposed loop detect method with ORB-SLAM2 and the random walk descriptor-based graph matching in \cite{lin2021topology}.
The authors of \cite{lin2021topology} have not released its source code before the submission.
To have a fair comparison, we faithfully reimplemented \cite{lin2021topology} on our own.
In our experiment, we set the number of random walks to 200 and the walk depth to 4.
The other parameters are consistent with the paper.
The element of the random walk descriptor consists of the semantic label and embedding of the object.
Table \ref{tb:loop_detection_results} shows the results of loop closure detection on TUM RGBD sequences.
The results show that, compared with ORB-SLAM2, our method can achieve 100\% accuracy and yield more true positives on these three sequences, although it detects fewer loop closure candidates due to its stricter graph matching.
Compared to the random walk descriptor-based method, our method achieves a higher recall and accuracy.

\begin{table}[h!]
    \centering
    \caption{Loop detection results on TUM RGBD dataset.}
    \label{tb:loop_detection_results}
    \begin{tabular}{|c|c|c|c|c|}
        \hline
        \noalign{\smallskip}
        \textbf{Sequence} & \textbf{Metric}    & \textbf{Ours} & \textbf{ORB-SLAM2} & \textbf{\cite{lin2021topology}} \\
        \noalign{\smallskip}
        \hline
                          & Detections         & 21            & 176                & 16                              \\
        fr1 room          & True Positives     & 21            & \textbf{26}        & 16                              \\
                          & False Positives    & 0             & 150                & 0                               \\
                          & After Verification & \textbf{21}   & 0                  & 16                              \\
        \hline
                          & Detections         & 44            & 188                & 35                              \\
        fr2 desk          & True Positives     & \textbf{44}   & 38                 & 35                              \\
                          & False Positives    & 0             & 150                & 0                               \\
                          & After Verification & \textbf{44}   & 1                  & 35                              \\
        \hline
                          & Detections         & 76            & 250                & 59                              \\
        fr3 office        & True Positives     & \textbf{76}   & 34                 & 51                              \\
                          & False Positives    & 0             & 216                & 8                               \\
                          & After Verification & \textbf{76}   & 1                  & 51                              \\
        \noalign{\smallskip}
        \hline
    \end{tabular}
\end{table}

\subsection{Loop Closure Results}\label{subsec:Loop_Closure_Results}
\begin{figure*}[htb]
    \centering
    \subfigure[Loop closure based on Superglue feature matching]{
        \begin{minipage}[b]{0.45\linewidth}
            \includegraphics[width=1\linewidth]{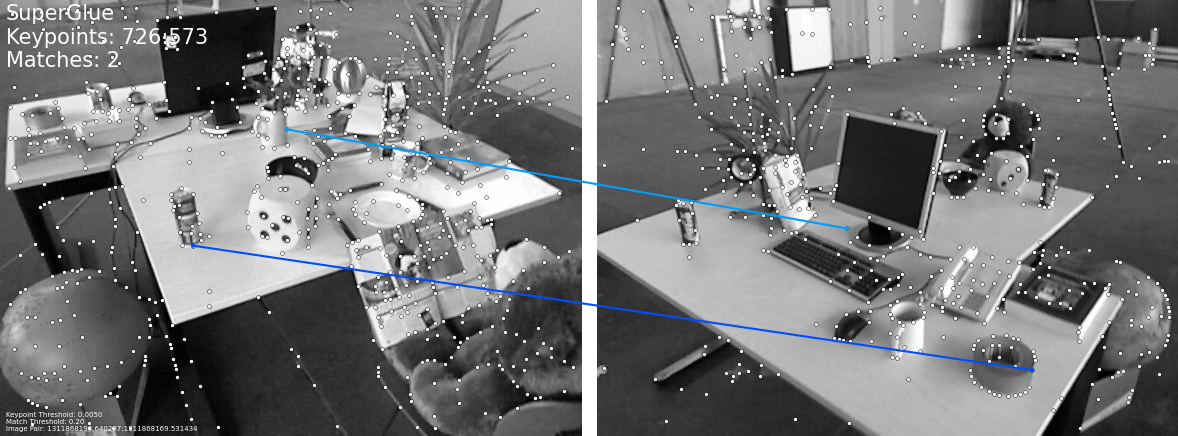}
            \includegraphics[width=1\linewidth]{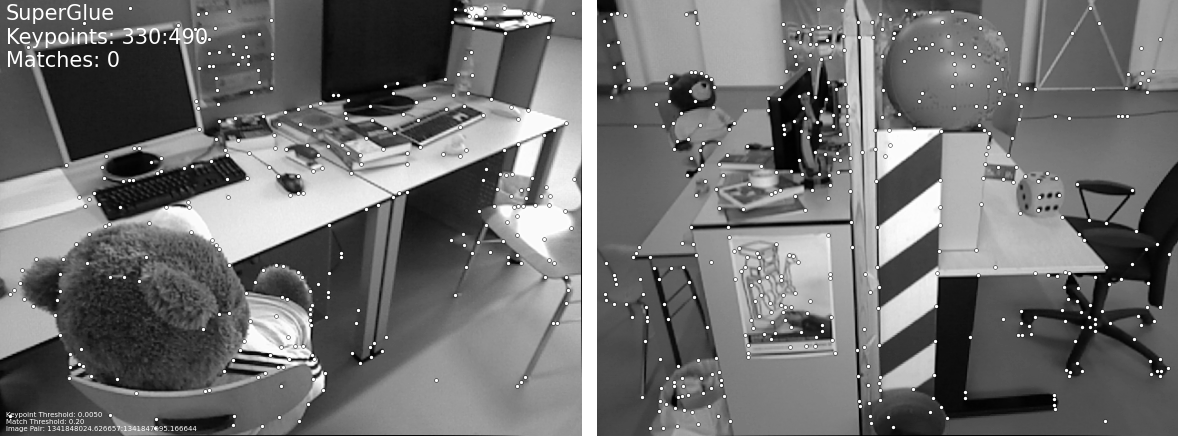}
        \end{minipage}}
    \subfigure[Loop closure based on our semantic graphs]{
        \begin{minipage}[b]{0.45\linewidth}
            \includegraphics[width=1\linewidth]{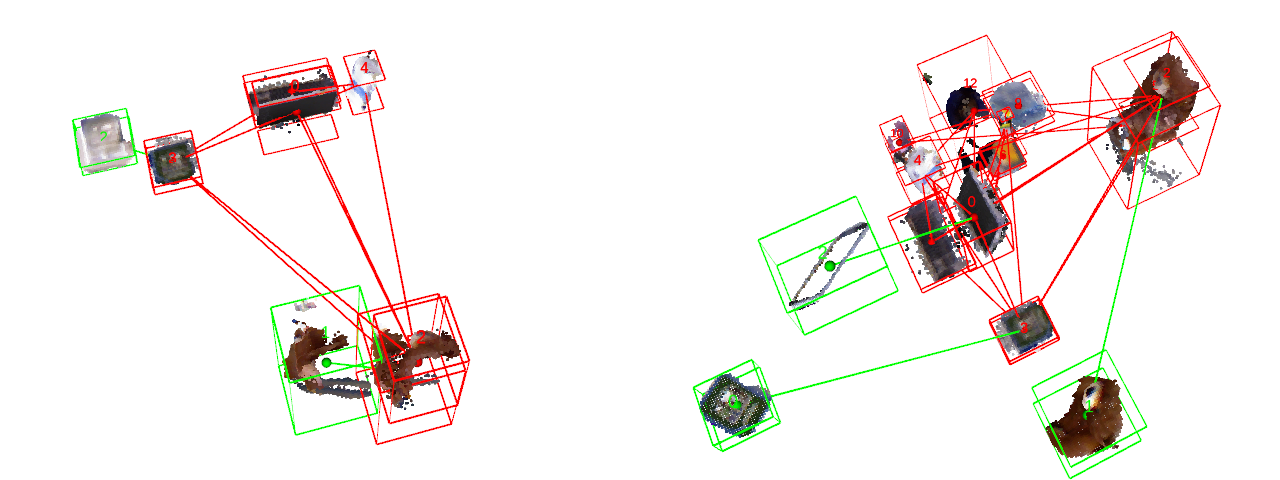}
            \includegraphics[width=1\linewidth]{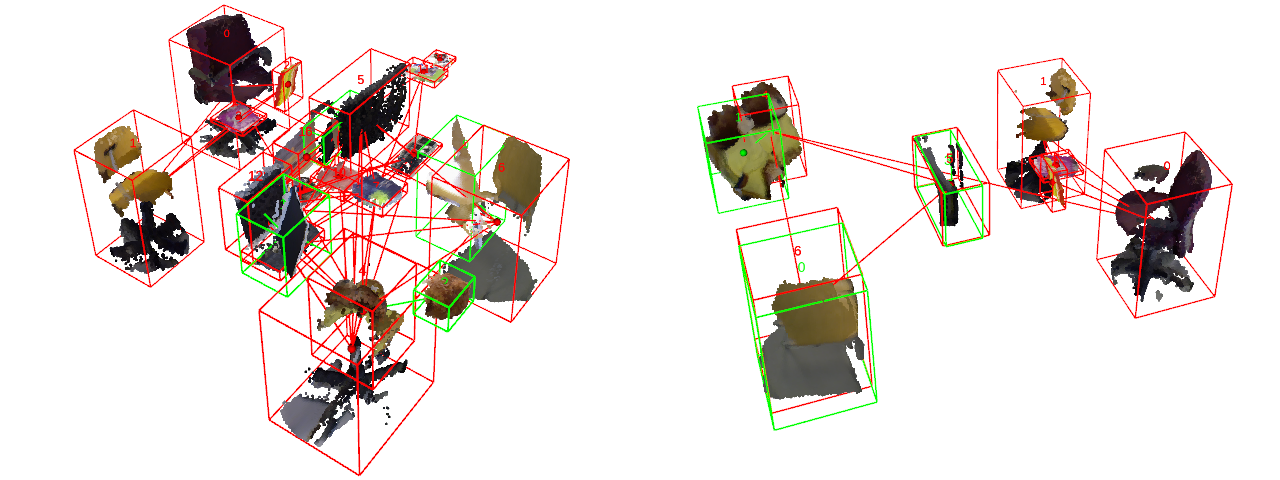}
        \end{minipage}}
    \caption{Two examples of challenging cases (significant viewpoint differences) in loop closure on the public TUM RGBD dataset. (a) Failure matching results based on Superglue. (b) Successful matching results based on our semantic graph matching. Green bounding boxes are objects in the local map. Red bounding boxes are objects in the global map. Green lines represent the correspondences of the objects. Red lines represent the co-visibility relationships.}
    \label{fig:loop_detection_results}
\end{figure*}

Figure~\ref{fig:loop_detection_results} shows two challenging scenes on the TUM RGBD dataset.
The first column is the result of Superglue \cite{sarlin2020superglue} feature matching.
The second column is the result of semantic graph matching.
Results show that the state-of-the-art learning-based matching method can not effectively perform loop closures in these challenging cases.
However, our semantic graph matching-based method can associate measurements from different viewpoints to the
object landmarks in the map and thus is highly robust to significant viewpoint differences.

\begin{figure*}[htb]
    \centering
    \subfigure[Failure cases of random walk descriptor-based graph matching in \cite{lin2021topology}.
        (Left) Random walks tend to lose information when there are many objects in the global map. As a result, two books (highlighted in the red circle) in the local map failed to match.
        (Right) Random walk descriptor-based graph matching often suffers false alarms when duplicated objects with similar topologies are in the global map. The monitor in the local map is incorrectly associated (highlighted with the green arrow) with the other monitor with id=5 in the global map.]{
        \begin{minipage}[b]{0.90\linewidth}
            \includegraphics[width=0.225\linewidth]{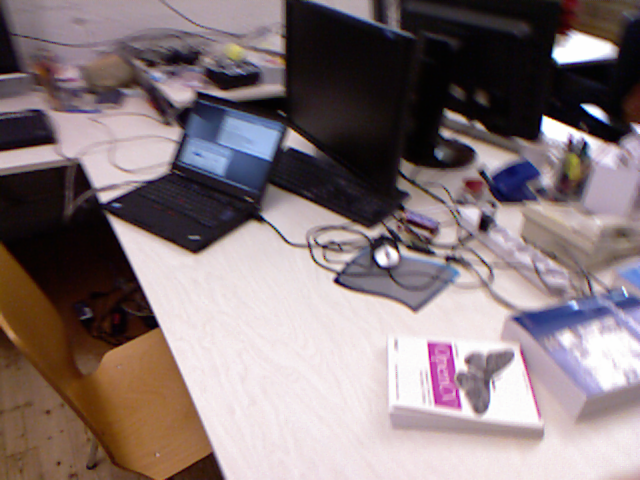}
            \includegraphics[width=0.225\linewidth]{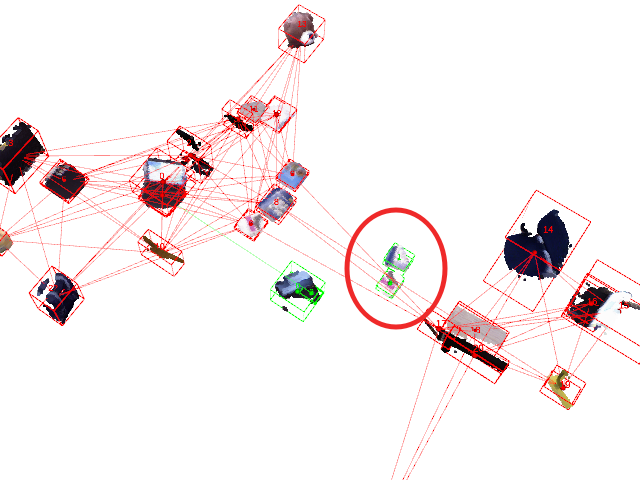}
            \includegraphics[width=0.225\linewidth]{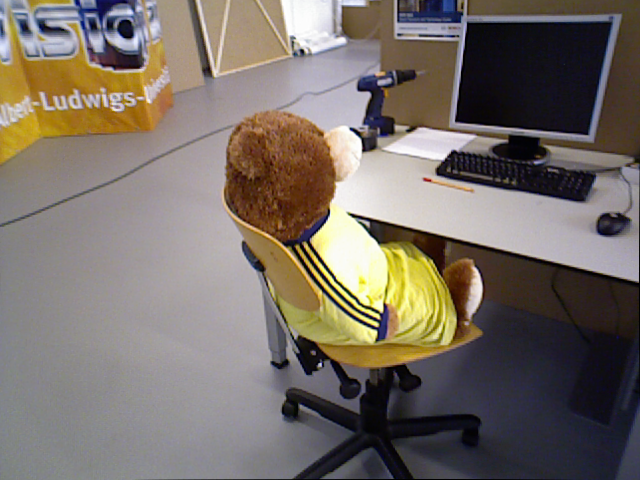}
            \includegraphics[width=0.225\linewidth]{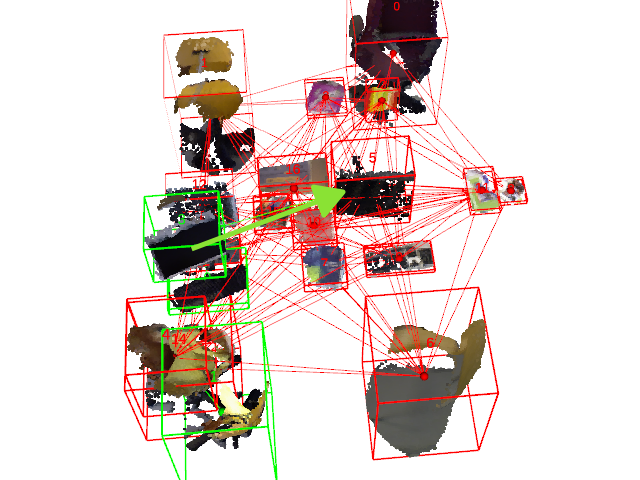}
        \end{minipage}}
    \subfigure[Failure cases of our method.
        (Left) The partially reconstructed desk is incorrectly associated (highlighted with the red arrow) because its center is closer to the desk with id=304 than the desk with id=275.
        (Right) The objects with id=498044 (keyboard), 509336(monitor), 454702(keyboard) have the same layout as objects with id=469041 (keyboard), 483912(monitor), 454702(keyboard). Thus, two objects are incorrectly associated (highlighted with red arrows).]{
        \begin{minipage}[b]{0.90\linewidth}
            \includegraphics[width=0.225\linewidth]{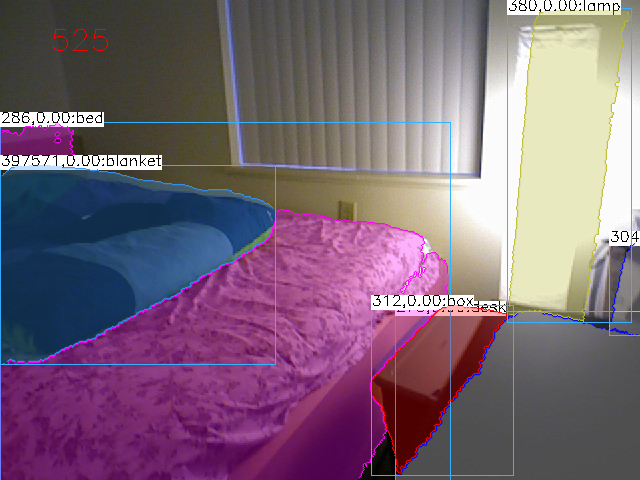}
            \includegraphics[width=0.225\linewidth]{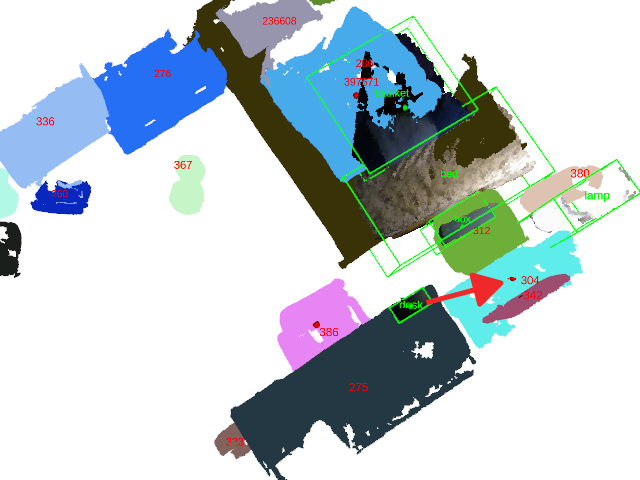}
            \includegraphics[width=0.225\linewidth]{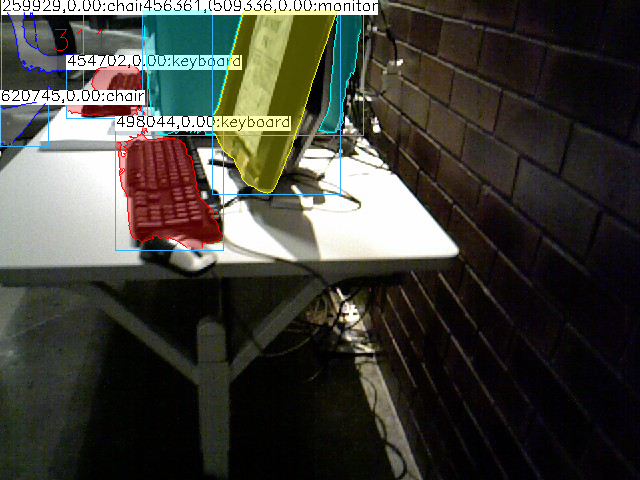}
            \includegraphics[width=0.225\linewidth]{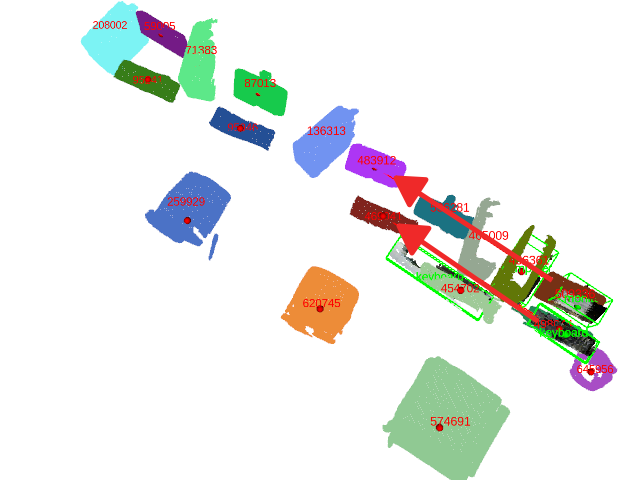}
        \end{minipage}}
    \caption{Four examples of failure cases. (a) Failure cases of random walk descriptor-based graph matching. (b) Failure cases of our method.}
    \label{fig:loop_detection_failures}
\end{figure*}

Figure~\ref{fig:loop_detection_failures} shows four false loop detections on the TUM RGBD and SceneNN datasets.
The first row is the failure cases of random walk descriptor-based graph matching in \cite{lin2021topology}, but successful based on our method. The reasons are two folded: firstly, random walks tend to lose information when there are many objects in the global map, and secondly, random walk descriptor-based graph matching often suffers falses alarms when duplicated objects with similar topologies are in the global map. The above two reasons explain why Table \ref{tb:loop_detection_results} shows that our method can achieve better recall and accuracy than the random walk descriptor-based method.
The second row is the failure cases of our method on the SceneNN dataset. Results show that partially reconstructed objects with inaccurate centers may cause false loop detections. Moreover, our method cannot handle scenes with the same object layout, e.g., a computer lab with many repeated object layouts. Combining feature points and semantics may solve this problem.
However, if there are multiple associated objects, we can eliminate these false loop detections by checking for topology and the spatial distance consistency between the matching objects in the two maps.

\subsection{Localization Performance}\label{subsec:Localization_Performance}
We evaluate the localization performance on the TUM RGBD and SceneNN datasets.
We compare our approach with ORB-SLAM2.
Table~\ref{tb:localization_results} shows the Root Mean Square translational Error (RMSE) of the trajectories. Note that we obtained the results of ORB-SLAM2 with the loop closing thread turned on. On fr2\_desk and fr3\_office sequences, our method is on par with ORB-SLAM2 since both methods detect enough loop closures. However, our method exhibits better localization performance on the other sequences as it can detect more challenging loop closures.


\begin{table}[h!]
    \centering
    \caption{Trajectory estimation mean error.}
    \label{tb:localization_results}
    \begin{tabular}{|c|c|c|c|}
        \hline
        \textbf{DataSet} & \textbf{Sequence} & \textbf{Ours}  & \textbf{ORB-SLAM2} \\
        \hline
                         & fr1 room          & \textbf{0.040} & 0.044              \\
        TUM RGBD         & fr2 desk          & \textbf{0.008} & 0.010              \\
                         & fr3 office        & 0.009          & \textbf{0.008}     \\
        \hline
                         & 021               & \textbf{0.066} & 0.106              \\
        SceneNN          & 025               & \textbf{0.086} & 0.116              \\
                         & 231               & \textbf{0.048} & 0.061              \\
        \hline
    \end{tabular}
\end{table}

\subsection{Runtime and Scalability}\label{subsec:runtime_scalability}
We evaluate the average running time of graph match on a Linux system with an Intel Core i7-7700K CPU at 4.20GHz. Table \ref{tb:loop_detection_avg_runtime} shows that our method is more than 2.5 times faster than the random walk descriptor-based graph matching method in \cite{lin2021topology}.

\begin{figure}[htb]
    \begin{center}
        \includegraphics[width=0.9\columnwidth]{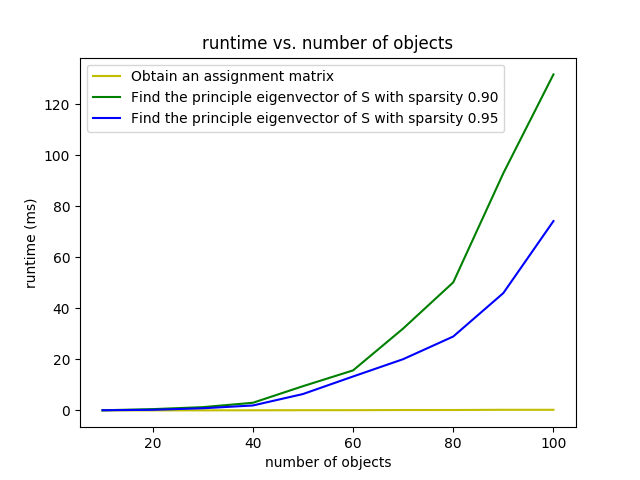}
    \end{center}
    \caption{Runing time for the graph matching steps varies with the number of objects and the sparsity of S.}
    \label{fig:scalability}
\end{figure}

Since the S matrix in equation $\left ( \ref{eqn:quadratic_assignment_formulation} \right ) $ is a highly sparse matrix (for the sequences in Table \ref{tb:loop_detection_avg_runtime}, the sparsity is larger than 0.95). The complexity of computing its principal eigenvectors is usually less than $O(n^{3/2})$, where $n=N \times M$. In our implementation, we call the Spectra library \cite{spectra_library}, which implements the Arnoldi/Lanczos method to find the principal eigenvectors of large symmetric sparse matrices efficiently. Figure \ref{fig:scalability} shows how the running time for the graph match steps varies with the number of objects (we set $N=M=$ number of objects in our experiments) and the sparsity of matrix $S$. The results show that our method scales well as the number of objects increases. For huge maps (containing thousands of objects), we can divide the huge map into several smaller submaps and then perform graph matching between the submaps.

\begin{table}[h!]
    \centering
    \caption{Average running time of graph match.}
    \label{tb:loop_detection_avg_runtime}
    \begin{tabular}{|c|c|c|c|}
        \hline
        \noalign{\smallskip}
        \textbf{Sequence} & \textbf{Ours (ms)} & \textbf{\cite{lin2021topology} (ms)} & \textbf{Sparsity of S} \\
        \noalign{\smallskip}
        \hline
        fr1 room          & \textbf{0.39}      & 0.99                                 & 0.98                   \\
        \hline
        fr2 desk          & \textbf{0.035}     & 0.15                                 & 0.96                   \\
        \hline
        fr3 office        & \textbf{0.14}      & 0.37                                 & 0.95                   \\
        \noalign{\smallskip}
        \hline
    \end{tabular}
\end{table}

\section{Conclusion}\label{sec:conclusion}
We have proposed a novel object-level data association to reconstruct the environment as 3D semantic maps. Then we perform loop closure based on semantic graph matching and object alignment. Finally, we jointly optimize camera trajectories and object poses in an object-level pose graph formulation. We have evaluated our methods on public TUM RGBD and SceneNN datasets. Experimental results demonstrate the effectiveness of the algorithms.

We believe that our method can further address the long-term localization challenges of robots, allowing robots to perceive the world in a more human-like manner. However, we need to address several limitations in future work. The partially reconstructed objects may affect the accuracy and robustness of semantic graph matching and object alignment. We plan to introduce learned representations to provide shape priors for better object reconstruction. Currently, our work only exploits semantic labels, spatial distances, and co-visibility. We plan to expand our algorithm with 6-DoF poses and 3D scene graphs.


\bibliographystyle{IEEEtran.bst}
\bibliography{references}

\end{document}